\documentclass[conference]{IEEEtran}
\IEEEoverridecommandlockouts
\usepackage{cite}
\usepackage{amsmath,amssymb,amsfonts}
\usepackage{algorithmic}
\usepackage{graphicx}
\usepackage{textcomp}
\usepackage{xcolor}
\usepackage{multirow}
\usepackage{tablefootnote}

\def\BibTeX{{\rm B\kern-.05em{\sc i\kern-.025em b}\kern-.08em
    T\kern-.1667em\lower.7ex\hbox{E}\kern-.125emX}}

\makeatletter % changes the catcode of @ to 11
\newcommand{\linebreakand}{%
  \end{@IEEEauthorhalign}
  \hfill\mbox{}\par
  \mbox{}\hfill\begin{@IEEEauthorhalign}
}
\makeatother % changes the catcode of @ back to 12
    
\begin{document}

\title{Automated Assessment of Critical View of Safety in Laparoscopic Cholecystectomy\\
}

\author{
\IEEEauthorblockN{Yunfan Li}
\IEEEauthorblockA{
\textit{Department of Computer Science}\\
\textit{Stony Brook University}\\
Stony Brook, USA \\
yunfli@cs.stonybrook.edu \\} 
\and
\IEEEauthorblockN{Himanshu Gupta}
\IEEEauthorblockA{ 
\textit{Department of Computer Science}\\
\textit{Stony Brook University}\\
Stony Brook, USA \\
hgupta@cs.stonybrook.edu}
\and
\IEEEauthorblockN{Haibin Ling}
\IEEEauthorblockA{ 
\textit{Department of Computer Science}\\
\textit{Stony Brook University}\\
Stony Brook, USA \\
hling@cs.stonybrook.edu}
\and
\IEEEauthorblockN{IV Ramakrishnan}
\IEEEauthorblockA{ 
\textit{Department of Computer Science}\\
\textit{Stony Brook University}\\
Stony Brook, USA \\
ram@cs.stonybrook.edu}
\and
\IEEEauthorblockN{Prateek Prasanna}
\IEEEauthorblockA{ 
\textit{Department of Biomedical Informatics}\\
\textit{Stony Brook University}\\
Stony Brook, USA \\
prateek.prasanna@stonybrook.edu}
\and
\IEEEauthorblockN{Georgios Georgakis}
\IEEEauthorblockA{
\textit{Department of Surgery}\\
\textit{Stony Brook University Hospital}\\
Stony Brook, USA \\
georgios.georgakis@stonybrookmedicine.edu}
\and
\IEEEauthorblockN{Aaron Sasson}
\IEEEauthorblockA{
\textit{Department of Surgery}\\
\textit{Stony Brook University Hospital}\\
Stony Brook, USA \\
aaron.sasson@stonybrookmedicine.edu}
}
\maketitle
\newcommand{\para}[1]{\noindent {\bf #1}}
\newcommand{\softpara}[1]{\noindent \underline{#1}}
\newcommand{\cbl}{\color{blue}}
\newcommand{\cb}{\color{black}}
\newcommand{\red}[1]{\textcolor{red}{#1}}
\newcommand{\blue}[1]{\textcolor{green}{#1}}
\newcommand{\green}[1]{\textcolor{green}{#1}}
\newcommand{\magenta}[1]{\textcolor{magenta}{#1}}
\newcommand{\eat}[1]{ }
\newcommand{\PP}[1]{\textcolor{cyan}{[PP: #1]}}

\newcounter{packednmbr}
\newenvironment{packedenumerate}{
\begin{list}{\thepackednmbr.}{\usecounter{packednmbr}
\setlength{\itemsep}{1pt}
\setlength{\labelwidth}{8pt}
\setlength{\leftmargin}{30pt}
\setlength{\labelsep}{4pt}
\setlength{\listparindent}{\parindent}
\setlength{\parsep}{1pt}
\setlength{\topsep}{3pt}}}{\end{list}}

\begin{abstract}
Cholecystectomy (gallbladder removal) is one of the most common procedures in the US, with more than 1.2M procedures annually.
%%%%%%
Compared with classical open cholecystectomy, laparoscopic cholecystectomy (LC) is associated with  significantly shorter recovery period, and hence is the preferred method.
However, LC is also associated with an increase in 
bile duct injuries (BDIs), resulting  in significant morbidity and mortality. 
%%%%%%%%%%%%%
The  primary cause of BDIs from LCs is misidentification of the cystic duct with the bile duct.
Critical view of safety (CVS) is the most effective of safety protocols, which is said to be achieved during the surgery if certain criteria are met. However, due to suboptimal understanding and implementation of CVS,
the BDI rates have remained stable over the last three decades.

In this paper, we develop deep-learning techniques to automate the assessment of CVS in LCs.
%%%%%%%%%%%%%%%%%
An innovative aspect of our research is on developing specialized learning techniques by incorporating domain knowledge to compensate for the limited training data available in practice.
In particular, our CVS assessment process involves a fusion of 
two segmentation maps followed by an estimation of a certain region of interest based on anatomical structures close to the gallbladder, and then finally determination of 
each of the three CVS criteria via rule-based assessment of structural information.
%%%%%%%%%%%%%%%%%%%%%%%%%%%%%
We achieved a gain of over 11.8\% in mIoU on relevant classes with our two-stream semantic segmentation approach when compared to a single-model baseline, and 1.84\% in mIoU with our proposed Sobel loss function when compared to a Transformer-based baseline model. For CVS criteria, we achieved up to 16\% improvement and, 
for the overall CVS assessment, we achieved 5\% improvement in balanced accuracy compared to DeepCVS under the same experiment settings.

% Our developed techniques can be used to raise alerts in real-time
% to prevent any duct or structure from being injured/cut.
 \end{abstract}

% \begin{keywords}
% Laparoscopic Cholecystectomy, .
% \end{keywords}

% Automatic assessment of critical view of safety (CVS) in laparoscopic cholecystectomy surgeries is a uniquely challenging problem due to low level of color separation in the surgical view and the lack of open-source datasets. Current methods for this task rely on neural networks to directly predict CVS labels, which are not only lacking in explainability, but also require large amount of training data to reach a certain 
% level of generalizability. In this paper, we introduce a pipeline which utilizes deep learning method for perception-oriented tasks, in this case semantic segmentation, and incorporates human knowledge for assessing CVS conditions and final decision making, which leads to more explainable results.

\section{Introduction}
\label{sec:aims}

Cholecystectomy is one of the most common surgical procedures in the US, done to remove
an inflamed or infected gallbladder.
%%%%%
Majority of cholecystectomy procedures are now done as laparoscopic
cholecystectomy (LC), as they are associated with shorter recovery times.
However, LCs are also associated with an increased number of bile duct injuries (BDIs), 
which occur due to limited field of vision.
%%%%%%%
% BDI is a serious complication that can result in significant morbidity and mortality. \PP{Can we provide some statistics of BDI incidence and its impact?}
%%%%%%%%%
BDIs resulting from LCs may lead to 
serious complications which can even endanger the patient's life and safety~\cite{barbier2014long,bdi-consq},
while driving up the medical litigation~\cite{alkhaffaf201015}
and healthcare costs to over a billion dollars in the US alone~\cite{berci2013laparoscopic}.
%%%%%%%%%%%%
A safety protocol, termed as critical view of safety (CVS), has been developed and widely embraced 
over the years, with the goal of minimizing misidentification of ducts and thus reduce incidence of 
BDIs. In spite of many evidences of the effectiveness of CVS protocol, the incidence of BDIs has 
not decreased over the past decades; the main reason for this stems from the insufficient implementation
and understanding of CVS criteria by the surgeons~\cite{daly2016current}. 
%%%%%%%%%%%%%%%%%%%%%%%%%%%%%%%%%%%%%%%%%%%%%%%%%%
Thus, automation of the CVS attainment in LC surgeries 
can potentially reduce 
incidence of BDIs in LCs.

\para{Vision.}
Our long-term vision is to develop a AI-driven surgical aid that 
will prevent
BDIs by a combination of real-time CVS assessment during LC,
enforcement of related safety processes  
(e.g., identifying and guiding surgeons to bailout strategies~\cite{sages_2021}), and 
training of surgeons via video reviews to improve their understanding of CVS and LC surgeries. 
%%%%%%%%%%%%%%%%%%%%%%%%%%%%
As a step towards the above vision, in this paper, we focus on developing
a technique to assess CVS based on its three criteria; such a technique 
can be used to raise alerts in real-time (i.e., while LC surgery is in progress) if an attempt is made to clamp or cut any
structure before a true CVS has been attained and thus, prevent BDIs.
%%%%%%%%%%%%%%%%%%%%
The key challenge in CVS assessment from learning techniques is 
the lack of sufficient training data (at most a few hundred LC surgery 
videos) as well as the intrinsic difficulties in CVS
assessment, such as the cluttered texture and occlusion among organs. 
Our approach addresses these
challenges by proposing a fusion approach followed by incorporation of clinical domain knowledge.
In particular, our approach involves estimating a region of interest based on 
anatomical structures around the gallbladder, and rule-based assessment of CVS criteria. We demonstrate that such an approach has a great potential in accurate detection of CVS by showing an advantage in performance on both individual CVS criteria and overall CVS classification when compared to CNN-based DeepCVS~\cite{mascagni2022artificial} as baseline.

% as to prevent any duct or other structure from being cut prior to attaining CVS.
% %%%%%%%%

% In recent years, various computer vision based methods have been proposed to potentially assist surgeons in identifying CVS~\cite{CITE}. However, to our knowledge, no prior works have proposed an explainable way to classify individual CVS conditions, and evaluations are limited on the frame level, which is not particularly significant in clinical setting.
% %%%%%%%%%%%%%%%%%
% In this paper, we propose a bottom-up approach for identifying critical view of safety on both the frame level and the video level.

\section{Background}
\label{sec:back}

In this section, we provide general background and related work.

\begin{figure}
    %\vspace{-0.2in}
    \centering
    \includegraphics[width=2in]{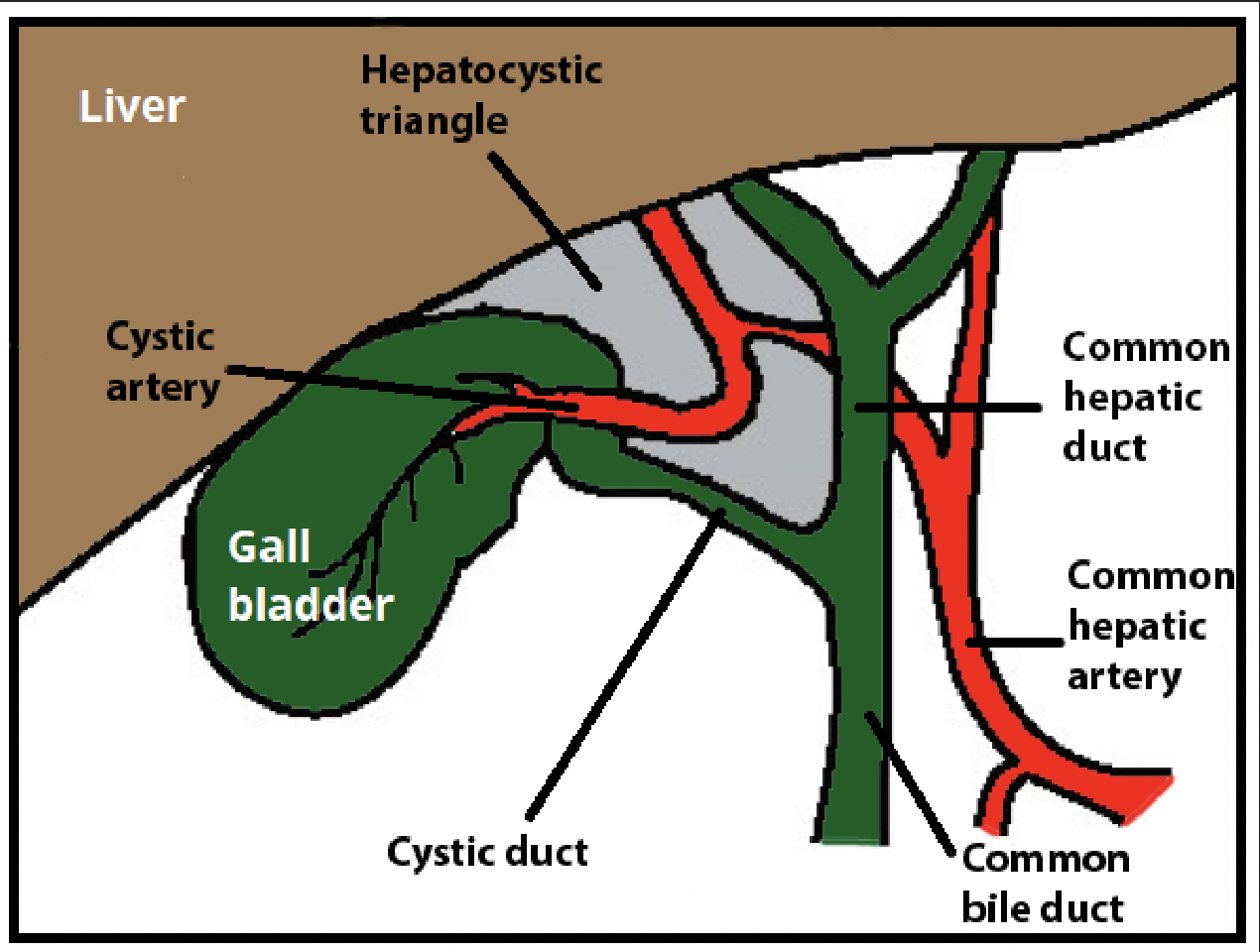}
   %  \vspace{-0.3in}
    \caption{Anatomy of hepatocystic triangle.~\cite{teachmeanatomy}}
    % \vspace{-0.2in}
    \label{fig:gb}
\end{figure}

\para{Laparoscopic Cholecystectomy (LC).}
Gallbladder is a small organ underneath the liver that concentrates and stores bile fluid. Inflammation and infection of the gallbladder may necessitate surgical removal
of the gallbladder, which is done via LC, a minimally invasive procedure  associated with quick recovery time. 
%%%%%%% 
LC, performed through four small incisions, uses a camera and surgical tools to remove the gallbladder. Removal of gallbladder essentially entails exposing (by removing the fat and fibrous tissues) and cutting the only two structures that connect it to the body: the cystic duct (CD) and the cystic artery.
% There are more than 1.2 million LCs performed annually\PP{This claim is not correct: 1.2 cholecystectomies are performed annually, not LCs}~\cite{million}.
%\footnote{Also, the gallbladder, being attached/glued to the liver bed, needs to be detached from it.}

\para{BDI Risks of LCs.}
The most feared adverse event of LC is bile duct injury (BDI), which occurs 
in thousands of cases in the US annually~\cite{bdi-consq}. BDIs largely
result from misidentification of the common bile duct as the cystic duct~\cite{strasberg2010rationale}, 
due to the increased complexity of LC procedures and limited field of vision. 
BDIs due to LCs may lead to 
serious complications and even endanger the patient's life and safety~\cite{barbier2014long,bdi-consq}. 
Overall, BDIs frequently result in a 3-fold increase in the 1-year mortality rate~\cite{tornqvist2012effect}, while driving up the medical litigation~\cite{alkhaffaf201015}
and healthcare costs to over a billion dollars in the US alone~\cite{berci2013laparoscopic,melton2002major,strasberg2017critical}.
% Finally, the  Society of American Gastrointestinal and Endoscopic Surgeons (SAGES) is raising awareness  of the safety of LCs through its SAGES safe cholecystectomy program~\cite{sages_2021}, underscoring the fundamental importance of 
% minimizing BDIs in LCs.

\para{The Critical View of Safety (CVS) Technique.}
Over the past few decades, surgeons have expended considerable effort in developing safe ways for identification of the cystic duct~\cite{kaczynski2015gallbladder}, of which the Critical View of Safety (CVS) technique is  considered to be the most effective at target identification and hence is widely embraced in LC procedures~\cite{sages_2021,cvs-gold}. 
CVS is said to be achieved if
the following three criteria are met:\footnote{CVS is a reworking
of the open cholecystectomy protocol
wherein the gallbladder is detached from 
the cystic plate (liver bed) 
so that it is attached to the body by only the two cystic structures which
can then be clipped. In laparoscopic surgery, as complete
separation of the gallbladder from the cystic plate makes
clipping of the structures difficult, we
require that only the lower part of the gallbladder
be separated~\cite{strasberg2010rationale}.}
\begin{packedenumerate}
\item[C1:] All fibrous and adipose
tissues cleared within the hepatocystic triangle (see Fig.~\ref{fig:gb}).
\item[C2:]
Separation
of the lower one-third of the gallbladder 
from the cystic plate (liver-bed).
\item[C3:] Two and only two
structures are seen to enter the gallbladder~\cite{strasberg2017perspective}.
\end{packedenumerate}

%\softpara{Impact of CVS Protocol.}
% The concept of the critical view of safety (CVS) 
% was introduced in 1995 as a way to unambiguously distinguish the biliary structures and thereby reduce  the incidence of 
% BDIs in LCs. 

\softpara{Impact and Limitation of CVS.}
The promise of CVS spurred several 
studies~\cite{yegiyants2008operative,nijssen2015complications}
on its effectiveness in the LC procedure, which provide strong evidence of the value of 
CVS as a means of unambiguously identifying biliary structures in LC.
% E.g., In a large single-institution
% series~\cite{cvs-gold,yeg-abs}, the observed BDI rate drops from 1/9 to 1/15, representing an order-of-magnitude of improvement in the safety of LC.
% Not surprisingly, CVS is now regarded as the gold standard for performing  safe LCs~\cite{cvs-gold,sages_2021}.
%
However, despite the  evidence of the efficacy of CVS in reducing mis-identification of CD,
BDI rates over the last 3 decades have remained  stable at 0.36\%–1.5\%~\cite{tornqvist2012effect}.
%%%%%%%%%
The primary reasons for this status quo are: 
insufficient or inadequate implementation of CVS~\cite{way2003causes}, and
weak understanding of CVS among many surgeons~\cite{daly2016current,chen2017increasing}.
Sometimes, overconfidence (partly due to the low incidence of BDIs) 
with LC also plays a part~\cite{daly2016current,nijssen2015complications,rawlings2010single,stefanidis2017often}.
Thus, automated assessment of CVS criteria has the potential to reduce BDIs, 
especially with the advances and contributions of computer vision in medical image analysis
over the recent years.

\para{Related Work.}
There have been two very-recent works on assessment of CVS.
%%%%%%%%%%%%%%%%%%%%%%%%%
In particular, Mascagni et al.~\cite{mascagni2022artificial} utilizes the semantic segmentation results of 
DeepLabV3+~\cite{chen2018encoder} and predicts binary labels of CVS criteria and overall CVS achievement from a compactly-designed CNN.
More recently, Murali et al.~\cite{murali2022latent} proposed incorporating graph neural networks (GNNs) to encode the latent scene graph in LC video frames, and shows improved performance over DeepCVS. However, these methods do not involve domain knowledge on CVS criteria and thus their results could not be easily analyzed or explained.
In another related work, Madani et al.~\cite{madani2022artificial} proposed using CNN-based semantic segmentation methods to 
identify safe and dangerous zones of dissections, which could serve as an important intermediary stage for CVS assessment. 

\section{Methodology}
\begin{figure*}[t]
  {\includegraphics[width=\linewidth]{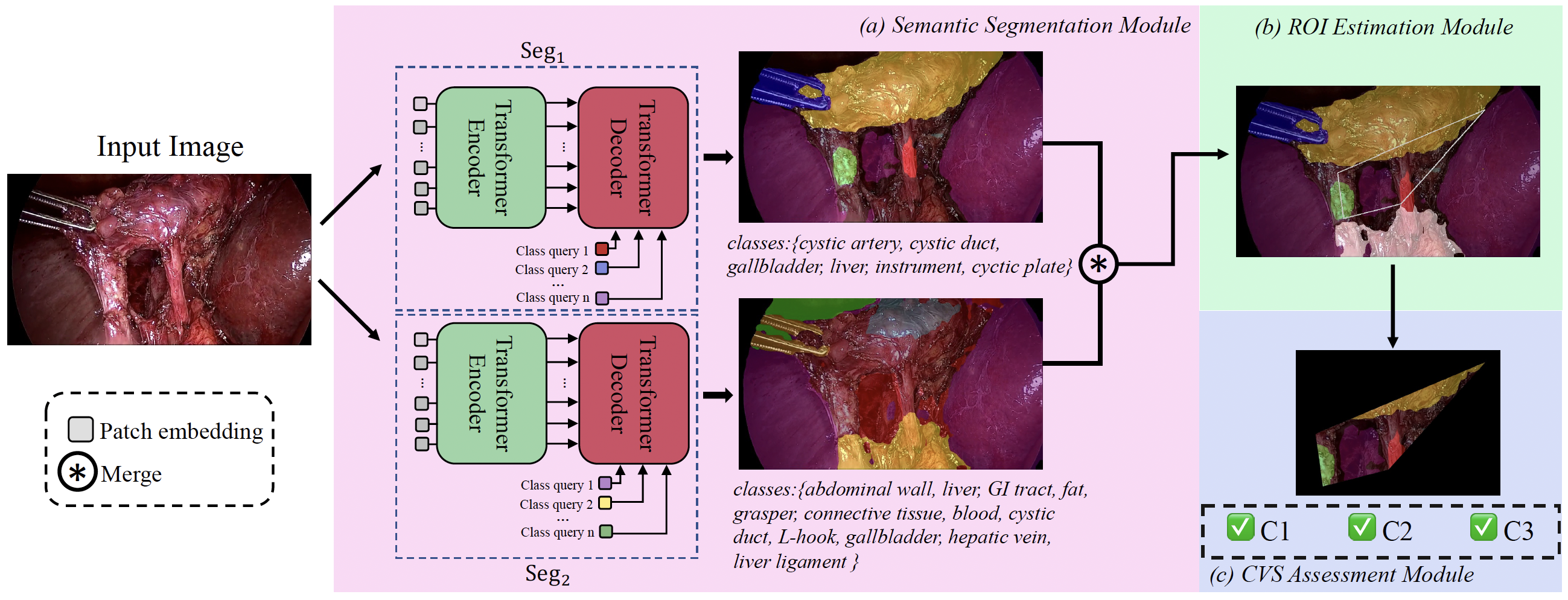} 
    \caption{Overall pipeline of our approach. The input frame is first segmented by two Transformer-based models. The segmentation maps are then merged for ROI estimation.  Finally, CVS conditions are evaluated based on ROI and segmentation maps.}
    \label{fig:pipeline}
  }
\end{figure*}

\para{Key Challenges in Automated CVS Assessment.}
Since the BDI incidence rate in LCs is extremely low (0.36\% to 1.5\%)~\cite{tornqvist2012effect}, 
a CVS detection technique must necessarily have very high accuracy (e.g., 90\% or more) 
to lower this BDI rate even further. Due to limited training data 
available,\footnote{One can realistically expect to curate a few hundred or at most
a few thousand LC surgical videos; by contrast, highly accurate ML models tend to use  millions of training samples.} such a high
accuracy is infeasible by direct application of machine-learning techniques, as seen in some
of the prior works. One approach to achieve such accuracy would be to
integrate extensive clinical/domain knowledge, as incorporating such knowledge has been shown to boost the accuracy of ML algorithms (e.g., \cite{xie2021survey,pape2019leveraging,konwer2022temporal}).
However, leveraging clinical domain knowledge in ML models can be quite challenging. 

\para{Method Pipeline and Key Contributions.}
Our approach tackles the aforementioned challenges by incorporating domain knowledge with limited training data.
In particular, our approach's pipeline is as follows (see Fig.~\ref{fig:pipeline}). First, to address the imbalance of classes in available datasets,
we segment each image frame by using two Transformer-based models trained on separate semantic segmentation datasets; relevant
classes from these two segmentation maps are then appropriately fused.
%%%%%%
Then, we use structural anatomic knowledge of the gallbladder and surrounding 
structures to estimate
the region of interest (ROI), which is used to efficiently assess the CVS conditions.
%%%%%
Finally, we assess each of the three CVS conditions based on their 
structural definitions, and then
the overall CVS as a conjunction of the three CVS conditions.
%%%%%%%%%%%%%
Overall, our main contributions include:
\begin{packedenumerate}
\item
Introducing a \textit{two-stream approach for semantic segmentation} to address the issue of class imbalance.
\item
Proposing a novel \textit{Sobel loss function} to reduce artifacts and over-segmentation around edges.
\item 
% Domain knowledge....talk about how ROI is defined, how segmentation is done, and how the CVS conditions are modified and assessed.
\textit{Integration of clinical domain knowledge:} Developing a rule-based approach for estimating ROIs and assessing CVS conditions in LC videos based on domain knowledge.
\end{packedenumerate}

% To implement our pipeline, we use \textit{CholecSeg8K} \cite{hong2020cholecseg8k} and \textit{CholecSeg170} dataset for the two segmentation models, and \textit{CVS6K} dataset for CVS assessment. Both \textit{CholecSeg170} and \textit{CVS6K} are constructed specifically for our method.

\subsection{Semantic Segmentation}
\label{sec:segmentation}

% Here, we discuss our semantic segmentation approach, which is a
% a crucial step for subsequent ROI estimation and CVS assessment. We first describe our two-stream framework.

% . CholecSeg8K is publicly available, already segmented -- so we must use it. BUT, it doesn't have CP + CA, and also CD is veyr low.
% . So, we created our own set of framkes whcih we call CholecSeg170 which takes care of the shortcomings -- but only 170 frames?.
% . WE could merge --- but we suspect not good performance. Shown in experiments.

% Thus, we created two independent segmentation models/maps. For common, we only use Seg 1. 
\para{Two-stream Segmentation and Fusion.}
For segmentation of LC frames, we wish to use the publicly available \textit{CholecSeg8K} dataset which includes 8,080 frames annotated with related classes. 
However, the \textit{CholecSeg8K} dataset is missing two important classes, viz.,
\textit{cystic plate} and \textit{cystic artery}, and has low number of pixels in \textit{cystic duct} class; all of these three classes are crucial to our approach (in particular, in 
estimation of the region of interest, discussed in the next section).
%%%%%%%%%%%%%%%%%%%%%%
To compensate for the above shortcomings, we created the \textit{CholecSeg170} dataset 
which includes annotations for cystic plate and cystic artery, and much higher proportion
of \textit{cystic duct} pixels.
%(see Section.~\ref{sec:dataset}).
%%%%%%%%%%%%%%%%%%%%%
We believe that training two separate segmentation models over the above two datasets
separately should yield better performance, especially on the important classes
\textit{cystic duct} and \textit{cystic artery}, than training a single segmentation model
over the union of the above datasets; our intuition is confirmed in our 
evaluation results (see Section.~\ref{sec:segmentation-result}). 

\begin{figure}[b]
\centering
  {\includegraphics[width=0.8\linewidth]{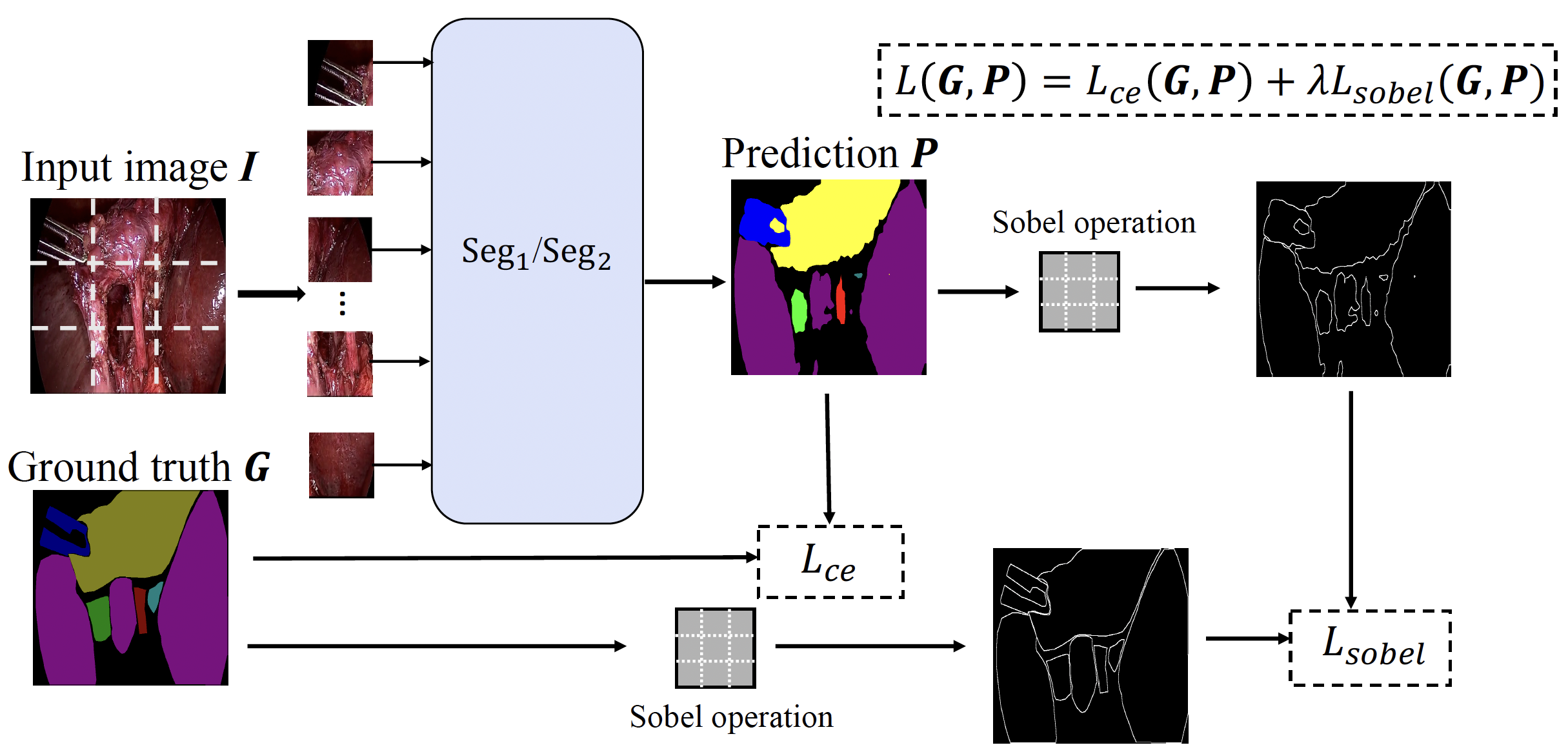} }
   \caption{Our proposed Sobel loss. It reduces artifacts and over-segmentation around edges by penalizing the difference between edge maps derived from segmentation maps.}
 \label{fig:segmentation-pipeline}
  % \vspace{-0.1in}
\end{figure}

Thus, the first segmentation model $\mathbf{Seg_1}$ is trained on the \textit{CholecSeg170} dataset, while the second  model $\mathbf{Seg_2}$ is trained on the \textit{CholecSeg8K} dataset. We use $\mathbf{Seg_1}$ for segmentation of 6 classes: \textit{cystic artery, cystic duct, gallbladder, liver, instrument, cystic plate}, while $\mathbf{Seg_2}$ is used for segmentation of only the \textit{fat} class.
%%%%%%%%%%%%%%%%%%%%%%%
For an input image $\mathbf{I}$, let $P_1 = \mathbf{Seg_1
}(\mathbf{I})$, $P_2 = \mathbf{Seg_2}(\mathbf{I})$. Then, the merged segmentation map is constructed by $P_{merged} = P_1 \oplus \mathbf{Fat}(P_2)$, where $\mathbf{Fat}$ denotes creating a mask of the \textit{fat} class. 
%%%%%%%%%%%%%%%%%%%%%

% \begin{figure}[htbp]
%  % Caption and label go in the first argument and the figure contents
%  % go in the second argument
%   \floatconts
%   {fig:distribution}  
%    {\caption{Class distribution of relevant classes in \textit{CholecSeg8K} (left) and \textit{CholecSeg170} (right).}}
%   {\includegraphics[width=1.0\linewidth]{distribution.png}}
% \end{figure}
%%%%%%%%%%%%%%%%%%%%%%%

\para{Sobel Loss Function.}
We use the Transformer-based Segmenter~\cite{strudel2021segmenter} model as the baseline for our semantic segmentation method. When evaluating the segmentation results, we observed that the edges between different anatomical classes are not clearly separated, causing artifacts and over-segmentation (see Section.~\ref{sec:segmentation-result}). To address this issue, we propose adding an edge-based constraint to the loss function. Specifically, we use the Sobel operator to generate class-agnostic edge information from the segmentation maps, and then apply Smooth \textit{L1} Loss~\cite{girshick2015fast} between the ground truth and predicted edges.

The Sobel operator uses of two $3\times3$ convolutional filters to calculate the approximations of the derivatives both vertically and horizontally. Given input image \textbf{I}, we calculate the gradient of the image \textit{Sobel}(\textbf{I}) as: $\textit{Sobel}(\textbf{I}) = \sqrt{G_{x}^2+G_{y}^2}$, where 
% \begin{equation}
%     \textit{Sobel}(\textbf{I}) = \sqrt{G_{x}^2+G_{y}^2}
% \end{equation}
% where
\begin{equation}
    G_x = \begin{bmatrix}
2 & 0 & -2 \\
4 & 0 & -4 \\
2 & 0 & -2 \\
\end{bmatrix}\ast \textbf{I},\qquad
G_y = \begin{bmatrix}
2 & 4 & 2 \\
0 & 0 & 0 \\
-2 & -4 & -2 \\
\end{bmatrix} \ast \textbf{I},
\end{equation}
$G_x$, $G_y$ are the two images containing horizontal and vertical derivatives respectively, and $\ast$ denotes the 2-D convolution operation.
 Given ground truth segmentation map $G$ and predicted segmentation map $P$, we define our Sobel loss function as:
\begin{equation}
    L_{Sobel}(G,P) = smooth_{L_1}(\textit{Sobel}(G)-\textit{Sobel}(P))
\end{equation}
where $smooth_{L_1}$ is the Smooth \textit{L1} Loss. Finally, our training objective is defined as 
\begin{equation}
    L(G,P) = L_{ce}(G,P) + \lambda L_{Sobel}(G,P)
\end{equation}
where $L_{ce}$ is the cross-entropy loss, and $\lambda$ is a hyperparameter. The segmentation model pipeline is shown in Fig.~\ref{fig:segmentation-pipeline}.

\subsection{Region of Interest (RoI) Estimation}
\label{section:roi-estimation}

In LC procedures, the assessment of CVS is mainly based on a specific region where the surgeon dissects tissue to expose cystic duct, cystic artery, and the cystic plate, and thereby
creating the CVS. In LC terminology, this region is referred to as the {\it hepatocystic triangle}. In most surgeries, the triangle is never fully visible since the 
surgeons usually only dissect to the point where cystic duct and cystic artery are 
sufficiently exposed while the common hepatic duct and common bile duct remain 
hidden.
%%%%%%%%%%%%%%%%%%%%%%%%%%%%%%%%%%%%%%%%%%%%%%%%%%%
Thus, in the LC surgery frames, we observe that only a part (in shape of a quadrilateral)
of the hepatoycstic triangle is visible. Hence, our region of interest (ROI) is of a quadrilateral shape with four sides.

\begin{figure}[t]
\centering
  {\includegraphics[width=\linewidth]{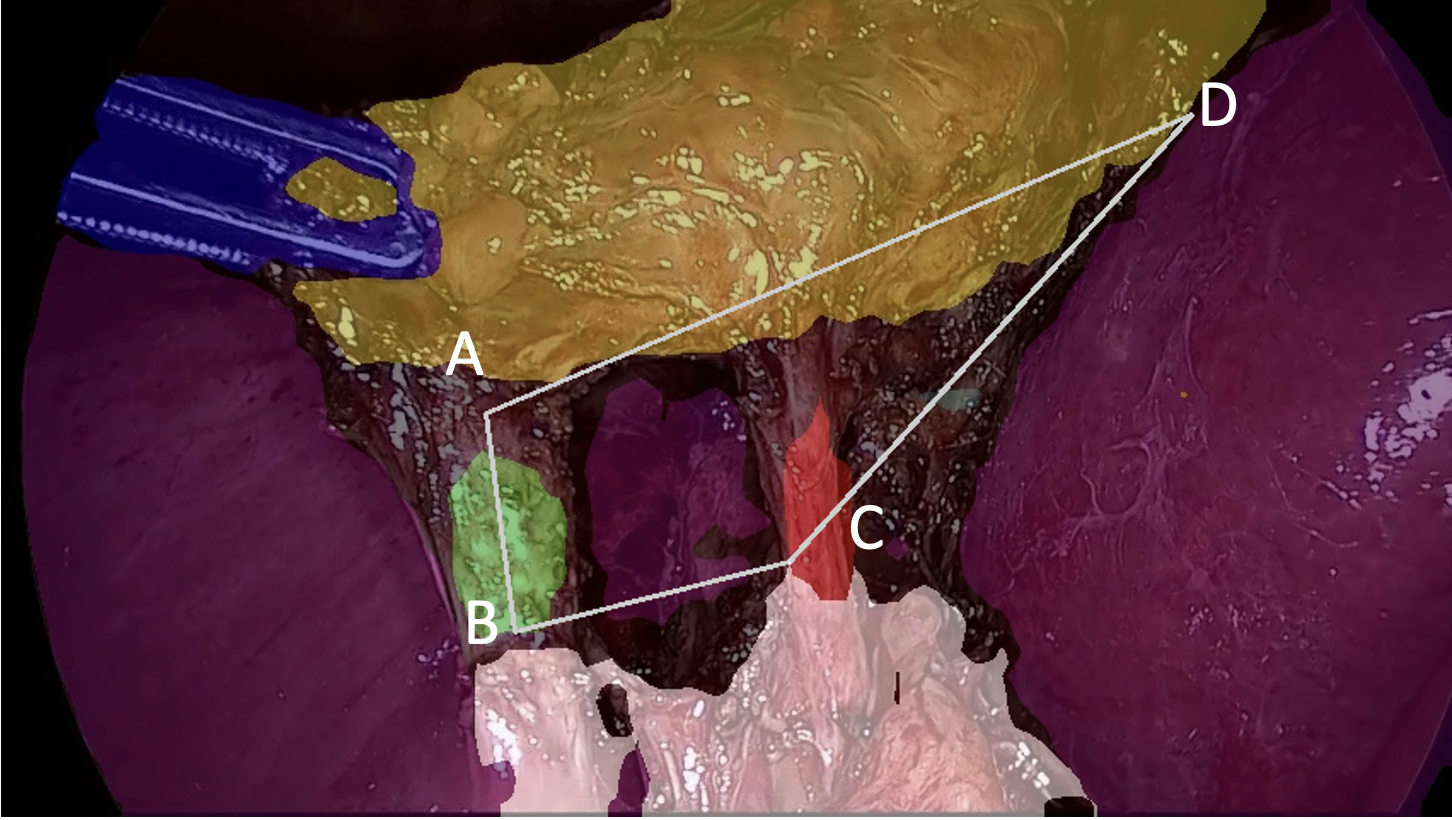}}
  \caption{ROI Quadrilateral.}
  \label{roi_points}
\end{figure}
The {\bf ROI quadrilateral} (see Fig.~\ref{roi_points}) is defined by anatomical structures 
around the gallbladder observed in the LC surgery videos. 
%%%%%%%%%%%%%%%%%%
Thus, we develop a clinically-motivated rule-based method to determine the ROI, 
rather than applying standard learning techniques as is typically done. 
%%%%
In particular, the ROI quadrilateral is formed by four points in an LC surgery 
image: (A) Cystic duct's end that is connected to the gallbladder; (B) Other end
of the (visible) cystic duct; (C) Intersection point between the liver edge and 
a line drawn from point B to the outline of the largest cluster 
of \textit{fat} class;
%tangent line of the fat region covering common hepatic duct drawn from point B;
(D) the point connecting the gallbladder to the liver. 
Note that the determination of point (C) is done to exclude the main cluster
of fat tissue from the ROI---we use the condition of such a 
quadrilateral being devoid of any fat tissue as
the sub-condition for the C1 criteria of CVS.

\begin{figure}[ht]
{\includegraphics[width=\linewidth]{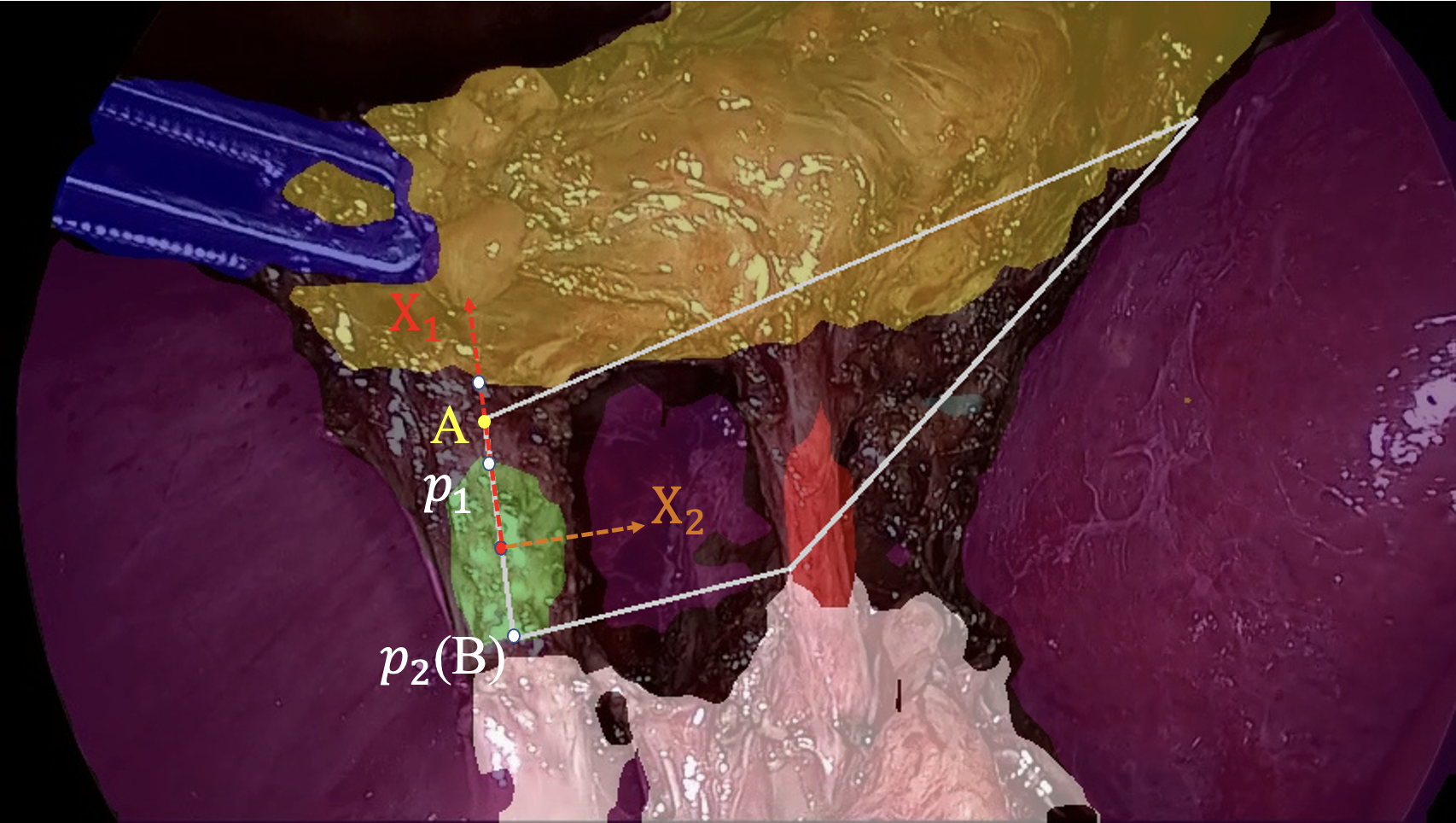} }
  \caption{Estimation of point \textbf{A} and \textbf{B} in our ROI estimation method. We first identify the two main components of the cystic duct $\mathbf{X_1}$,$\mathbf{X_2}$ cluster using PCA. Then we extend $\mathbf{X_1}$ in both directions from the centroid of the cluster to find $p_1$ and $p_2$. Finally, we define the mid-point between $p_1$ and its nearest neighbour on the gallbladder edge as \textbf{A}, and $p_2$ as \textbf{B}.}
  \label{fig:quad2}
\end{figure}

In a segmented frame, we estimate the above defined four points as follows. 
%%%%%%%%%%%%%
First, we estimate points $A$ and $B$ as follows (see Fig.~\ref{fig:quad2}) .
We perform principal component analysis (PCA) on the main cluster $\mathbf{C_{duct}}$ of 
\textit{cystic duct} pixels, as detected by the first segmentation model $\mathbf{Seg_1}$.
Let the two primary components obtained from PCA be $\mathbf{X_1}$ and $\mathbf{X_2}$, with 
$\mathbf{X_1}$ being the one with a higher angle (almost perpendicular) to the gallbladder edge. 
%%%%%%%%%%%%%%%%%
Next, we create a line segment 
by starting from the centroid of the cluster $\mathbf{C_{duct}}$ and extending
in both directions along $\mathbf{X_1}$ till the outline of the cluster is reached;
let the endpoints of this line segment be $p_1$ and $p_2$, with $p_1$ being the point closer
to the gallbladder. 
%%%%%%%%%%%%%%%
We define $A$ to the point between $p_1$ and its nearest neighbour on the gallbladder edge, 
and $B$ as $p_2$.
%%%%%%%%%%%%%%%%%%%
To estimate the point $C$, we start with the line connecting $A$ and $B$, and rotate it clockwise till 
it intersects with the main cluster of \textit{fat} tissue; the intersection point is assigned to be
point $C$.
%%%%%%%%%%%%
Finally, we estimate the point $D$ as follows. 
%%%%%%%%%%%%%%%%%%%%%%%%%%%%
Since the segmentation maps usually do not yield a unique point where the gallbladder and liver edges
intersect,  we choose a pair of points, one from each edge, that has the minimal Euclidean distance between them; for this, we use a modified KD-Tree Nearest Neighbour algorithm~\cite{maneewongvatana1999analysis}. The point $D$ is 
defined as the midpoint between these two points. 

\subsection{CVS Assessment}

Given the semantic segmentation maps and the ROI quadrilateral in an image frame,
we develop a rule-based method to determine attainment of each of the three CVS criteria and thus the CVS. Recall the three CVS conditions from Section.~\ref{sec:back}.
%%%%%%%%%%%%%%%%%%%%%%%%%
For \textbf{C1}, which is to check for fat or fibrous tissue in the hepatocystic
triangle (and thus, the ROI quadrilateral), we determine attainment of C1 condition
based on following two conditions: 
(a) No \textit{fat} pixels in the ROI; 
(b) The size of the cluster of \textit{liver} pixels in the ROI 
is more than a certain threshold $T_{liver}$. 
Note that the \textit{fat} and \textit{liver} classes are determined by
$\mathbf{Seg_2}$ and $\mathbf{Seg_1}$ segmentation maps respectively.
If both the above conditions are satisfied, we consider C1 condition to be satisfied.
%%%%%%%%%%%%%%%%%%%%%%%%
For \textbf{C2}, if the size of the cluster of \textit{cystic plate} pixels in the ROI surpasses 
a certain threshold $T_{cp}$, it is considered satisfied. 
%%%%%%%%%%%%%%%%%%%%%%%
For \textbf{C3}, if exactly one cluster of \textit{cystic duct} pixels and one cluster of \textit{cystic artery} pixels are detected by $\mathbf{Seg_1}$ in the ROI, it is considered satisfied. We empirically set $T_{liver}=100$ and $T_{cp}=100$ to eliminate some of the noisy predictions. 
%%%%%%%%%%%%

\section{Results}

In this section, we introduce the datasets we used for development and evaluation of our techniques and the results of our method.
\subsection{Datasets}
\label{sec:dataset}
\begin{figure}[b]
    \centering    
    \includegraphics[width=\linewidth]{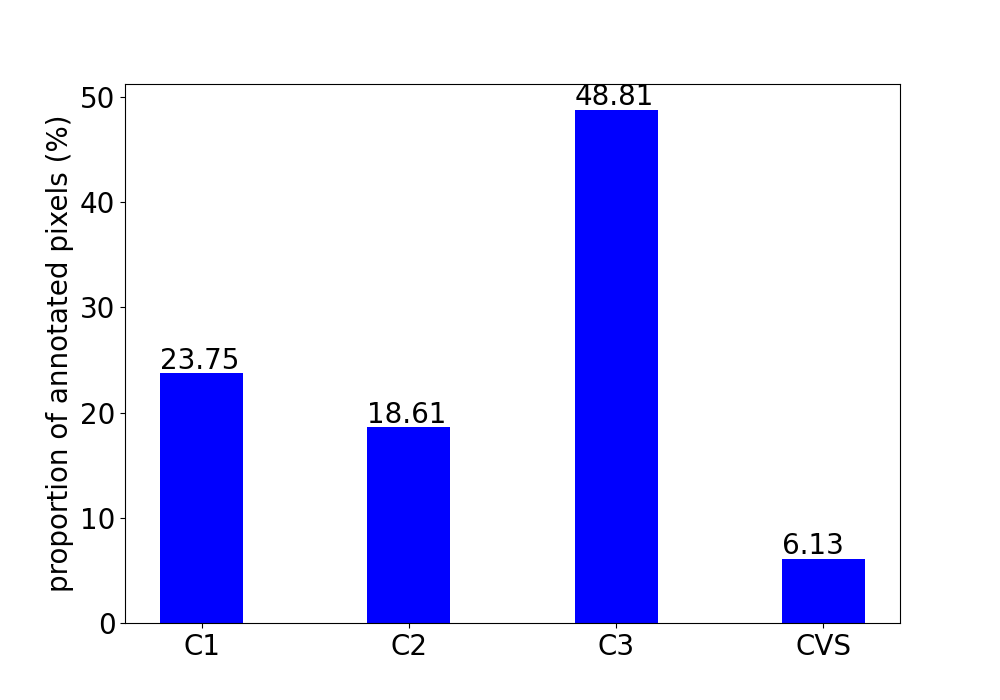}
    \caption{Proportion of positive examples in \textit{CVS6K}.}
    \label{fig-proportion}
\end{figure}

The combined \textit{Cholec80}~\cite{twinanda2016endonet}and \textit{m2cai16-workflow}~\cite{stauder2016tum} dataset consists of 117 videos after excluding duplicate cases~\cite{madani2022artificial}. We use the 17 videos from the \textit{CholecSeg8K} dataset as the development set and the remaining 100 as the evaluation set. The development set consists of two separate semantic segmentation datasets, namely \textit{CholecSeg8K} and \textit{CholecSeg170}. The evaluation set, named \textit{CVS6K}, consists of 6,000 frames with only binary CVS annotations.\\
\textbf{CholecSeg8K.} The \textit{CholecSeg8K} dataset is a publicly available semantic segmentation dataset based on the \textit{Cholec80} dataset. In total, 8,080 frames were collected from 17 videos in the \textit{Cholec80} dataset, and 13 different semantic classes (including background) were annotated. Most relevant classes in LC are annotated, such as \textit{liver, fat, gallbladder} and \textit{cystic duct}. However, \textit{CholecSeg8K} is highly unbalanced in class distribution, and some crucial classes for assessing CVS, such as \textit{cystic plate} and \textit{cystic artery}, are absent from the dataset.\\ 
\textbf{CholecSeg170.} To address the limitations of \textit{CholecSeg8K}, we collected 170 frames from the same 17 videos to form a separate semantic segmentation dataset, which we call the \textit{CholecSeg170} dataset. For each video, 10 frames are manually selected close to the \textit{ClippingCutting} stage as defined in \textit{Cholec80}, where most anatomical structures necessary for evaluating CVS are visible. The selected frames are annotated with the following 7 semantic classes: \{\textit{cystic artery, cystic duct, gallbladder, instrument, liver, cystic plate, background}\}. Additionally, ground truth CVS conditions are labeled for each frame.The 170 frames are divided into 140 frames for training and 30 frames for validation. \\
\textbf{CVS6K.} The 100 videos which are not included in the semantic segmentation datasets are used to construct the CVS evaluation set. We first sample a one minute clip at 1fps from each video, all of which near the \textit{ClippingCutting} stage of the videos, when CVS conditions can be clearly evaluated in most frames. For each frame, we assign three binary labels corresponding to the three criteria of CVS as suggested by SAGES~\cite{sages_2021}. If and only if all three criteria are satisfied in a frame do we consider CVS achieved in that frame.
%%%%%%%
% For each video clip of 60 frames, if the number of frames that satisfy a certain criterion is greater or equal to 1, we consider that criterion satisfied in the clip. If and only if all three criteria are satisfied in a clip do we consider CVS achieved in that clip. Binary labels of whether CVS is achieved are assigned at a video level accordingly. 
%%%%%%%%%%%%%%
The proportions of positive examples on the dataset is shown in Fig.~\ref{fig-proportion}.  All annotations on the CVS evaluation dataset are verified independently by two experienced oncology surgeons (co-authors).

\subsection{Semantic Segmentation}
\label{sec:segmentation-result}

\begin{figure*}[t!]
 \centering
  {\includegraphics[width=0.8\linewidth]{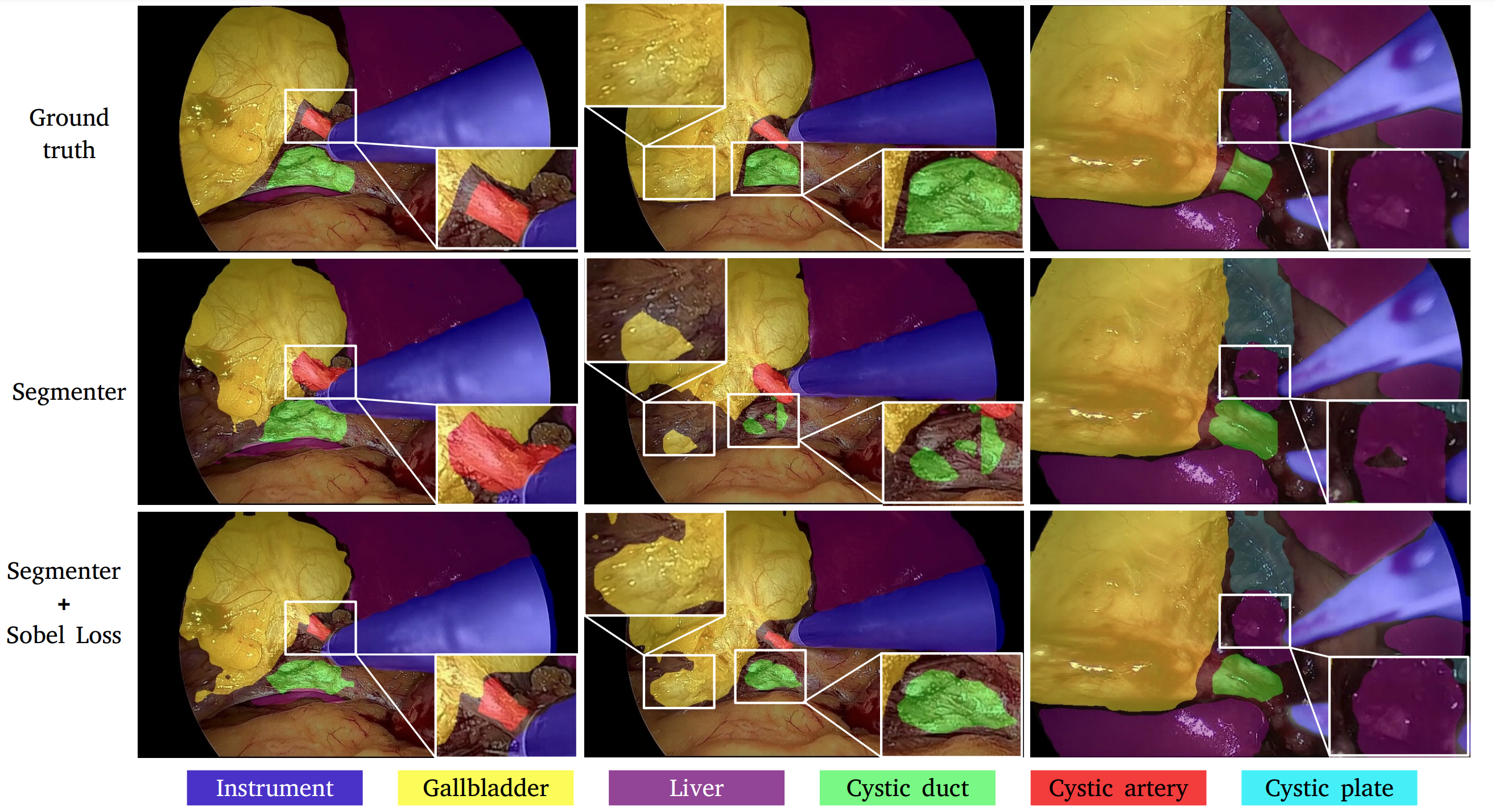}}
 \caption{Qualitative results. Our proposed Sobel Loss reduced over-segmentation of cystic artery in column 1, and improved on the artifacts/fragmented segmentations of gallbladder, cystic duct, and liver (columns 2, 3).}
  \label{fig:segmentation}
\end{figure*}

\begin{table*}[t]
    \centering
    \small
      \caption{Comparison of one vs.\ two-model segmentation approaches, in terms of  IoU.}
    \begin{tabular}{|c|c|c|c|c|c|c|}
    \hline
         {\bf Approach} & Gallbladder & Liver & Cystic Duct & Cystic artery & Cystic plate & Instrument\\
         \hline
         {\bf Single model} & 0.8964 & \textbf{0.9244} & 0.4978 &    0.0& 0.4229 & \textbf{0.8989} \\
         \hline
         {\bf Two-stream} & \textbf{0.9139} &  0.8913 & \textbf{0.6833} & \textbf{0.4484} & \textbf{0.5713} & 0.8433\\
         \hline
    \end{tabular}
  
    \label{tab:twostream}
\end{table*}

We start by evaluating the effectiveness of our two-stream segmentation approach by computing the IoU metric on each relevant class in TABLE~\ref{tab:twostream}. We observe
that the two-stream approach improves the IoU by 11.85\% on average, and the improvements are especially significant on low-frequency classes like \textit{cystic duct} (18.55\%), \textit{cystic artery} (44.84\%), and \textit{cystic plate} (14.84\%). We also assess the enhancement resulting from the proposed  Sobel loss on the validation set of {\it CholecSeg170} in TABLE~\ref{tab:segres}. We see that the Sobel loss function resulted in 1.84\% improvement in mIoU and 1.8\% improvement in Dice score compared to Segmenter baseline.
We used $\lambda=1$ when deploying Sobel loss.
%%%%%%%%%

We also evaluated {\bf qualitative results} in Fig.~\ref{fig:segmentation}. We see
that our proposed Sobel loss penalizes noisy predictions around edges, leading to more inter-class separation and thereby creating more defined edges on anatomical structures and organs. Additionally, it also reduces noisy patches often observed from the baseline model.

\begin{table}
    \centering
     \caption{Comparison of Sobel loss based segmentation and the baseline method.}
    \begin{tabular}{|c|c|c|c|}
    \hline
         Model/Metric & mIoU & Acc. & Dice\\
         \hline
         Baseline & 0.7270 & \textbf{0.9372} & 0.8247 \\
         \hline
         Baseline+Sobel loss & \textbf{0.7454} & 0.9323 & \textbf{0.8427} \\
         \hline
    \end{tabular}
   
    \label{tab:segres}
\end{table}

% \begin{figure}[htbp]
% \floatconts
%   {fig:roi}
%   {\caption{Qualitative results of region of interest (ROI) estimation.}}
%   {\includegraphics[width=1.0\linewidth]{ROI.png}}
% \end{figure}

% \subsection{ROI Estimation}
% We qualitatively evaluate the effectiveness of our ROI estimation method.

\subsection{CVS Conditions and CVS Assessment}
\label{sec:cvs-result}
We present the accuracy (Acc.), balanced accuracy (Bacc.), Positive Predictive Value (PPV) and Negative Predictive Value (NPP) on the independent \textit{CVS6K} dataset in TABLE~\ref{cvs1}. 
%%%%%%%%%%%%%%%%%%%%%%
For the baseline approach, we re-implemented DeepCVS according to the descriptions in~\cite{mascagni2022artificial}, with slight modification to fit our experiment settings, and for the purpose of fair comparison.
%%%%%%
In particular, we trained two separate DeepLabV3+ semantic segmentation models on \textit{CholecSeg170} and \textit{CholecSeg8K} datasets. The segmentation maps are fused the same way as described in Section.\ref{sec:segmentation}. The CNN for classification of CVS conditions are implemented according to~\cite{mascagni2022artificial} except for the first layer.
%%%%%%%%%%%%%%%%%%%%%%%%%
As may be observed in TABLE~\ref{cvs1}, our rule-based method significantly outperforms the baseline model on both independent CVS criteria and overall CVS assessment, and shows more consistent performance among different CVS conditions. 
\begin{table}
\small
    \centering
     \caption{Results of CVS assessment compared to DeepCVS.}
    \scalebox{0.85}{
  \begin{tabular}{|c|c|c|c|c|c|c|c|c|}\hline
    & \multicolumn{2}{|c|}{C1} 
    & \multicolumn{2}{|c|}{C2} &  \multicolumn{2}{|c|}{C3} & \multicolumn{2}{|c|}{CVS} \\
    \hline
    \multirow{4}{*}{DeepCVS}
    & Acc. & 0.72 & Acc. & 0.39 & Acc. & 0.54 & Acc. & 0.92 \\
    & Bacc. & 0.48 & Bacc. & 0.49 & Bacc. & 0.53 & Bacc. & 0.49 \\
    & PPV & 0.14 & PPV & 0.18 & PPV & 0.53 & PPV & NaN\footnotemark \\
    & NPV & 0.75 & NPV & 0.80 & NPV & 0.54 & NPV & 0.93\\
    % \hline
    % \multirow{3}{*}{CNN}
    % & Acc. & 0.746 & Acc. & 0.245 & Acc. & 0.482 & Acc. & \textbf{0.930} \\
    % & PPV & 0.204 & PPV & 0.189 & PPV & 0.468 & PPV & 0.070\\
    % & NPV & 0.761 & NPV & 0.854 & NPV & 0.494 & NPV & 0.938\\
    \hline
    \multirow{4}{*}{Ours}
    & Acc. & \textbf{0.76} & Acc.& \textbf{0.79} & Acc. & \textbf{0.69} & Acc. & 0.92 \\
    & Bacc. & \textbf{0.57} & Bacc. & \textbf{0.65} & Bacc. & \textbf{0.69} & Bacc. & \textbf{0.54}\\
    & PPV & \textbf{0.49} & PPV. & \textbf{0.43} & PPV & \textbf{0.72} & PPV & \textbf{0.23}\\
    & NPV & \textbf{0.79} & NPV & \textbf{0.86} & NPV & \textbf{0.67} & NPV & \textbf{0.94}\\

    \hline
          
    \end{tabular}

    }
    \label{cvs1}
\end{table}

\footnotetext{PPV is undefined in this case since all frames are predicted as negative in CVS.}

\section{Conclusion}

In this work, we have addressed a critical unmet clinical need, viz, assessing CVS in LC procedures to help minimize incidence of BDIs. 
% To overcome the challenge of limited training data, we develop techniques that leverage domain knowledge at various stages, and demonstrate that our developed techniques outperform baseline approaches both qualitatively and quantitatively.
We developed a 3-step pipeline, which addresses the issues of class imbalance and artifacts in semantic segmentation, while also incorporates domain knowledge for more accurate CVS assessment. The results show great promise in future applications in computer-assisted LC procedures. However, one limitation of our approach is that it heavily relies on the quality of the segmentation results and does not include a reasonable fail-safe mechanism when segmentation models produce undesirable results. To address this challenge, we aim to develop methods that take advantage of segmentation-failure detection techniques in our future work.

\section*{Acknowledgment}
We would like to acknowledge Twinanda et al.~\cite{twinanda2016endonet} and Hong et al.~\cite{hong2020cholecseg8k} for making their datasets publicly available to the research community.

Research reported in this publication was supported by National Science Foundation (NSF) under award numbers FET-2106447, CNS-2128187, 2153056, 2125147, 2113485, 2006655 and National Institutes of Health (NIH) under award numbers R01EY030085, R01HD097188, 1R21CA258493-01A1. The content is solely the responsibility of the authors and does not necessarily represent the official views of the NSF and the NIH.

\bibliographystyle{IEEEtran}

\bibliography{ref}

\end{document}